\def\year{2018}\relax
\documentclass[letterpaper]{article} %DO NOT CHANGE THIS
\usepackage{aaai18}  %Required
\usepackage{times}  %Required
\usepackage{helvet}  %Required
\usepackage{courier}  %Required
\usepackage{url}  %Required
\usepackage{graphicx}  %Required

\usepackage{bm}
\usepackage{algorithm2e}
\usepackage{subfigure}
\usepackage{epsfig}
\usepackage{epstopdf}
\usepackage{url}
\usepackage{multirow}
\usepackage{comment}
\usepackage{color}
\usepackage{makecell}
\usepackage{amssymb}
\usepackage{amsthm}
\usepackage{soul}
\usepackage{bm}
\usepackage{amsmath}

\begin{document}
\title{Robust Sparse Coding via Self-Paced Learning}
\author{Xiaodong Feng\\ Department of Information Management, University of Electronic Science and Technology of China\\
Chengdu, Sichuan, China\\
fengxduestc@126.com }
\maketitle
\begin{abstract}
Sparse coding (SC) is attracting more and more attention due to its comprehensive theoretical studies and its excellent performance in many signal processing applications. However, most existing sparse coding algorithms are nonconvex and are thus prone to becoming stuck into bad local minima, especially when there are outliers and noisy data.
To enhance the learning robustness, in this paper, we propose a unified framework named $Self$-$Paced~Sparse~Coding$ (SPSC), which
gradually include matrix elements into SC learning from easy to complex.
We also generalize the self-paced learning schema into different levels of dynamic selection on samples, features and elements respectively.
Experimental results on real-world data demonstrate the efficacy of the proposed algorithms.
\end{abstract}
\section{Introduction}
Learning effective representations for high dimensional data plays an essential role in machine learning tasks for various types of data, such as text, image, speech and video data.
Among these techniques, sparse coding is attracting more and more attention due to its comprehensive theoretical studies and its
excellent performance in many signal processing applications, especially computer vision tasks, including supervised image classification \cite{Ref:facerec}, semi-supervised image classification~\cite{Ref:lu2015noise}, image clustering \cite{Ref:Yang2014Data,Ref:feng2017multi}, image restoration \cite{Ref:Dong2015Image}, etc.
Usually, in these areas, the signals studied are image feature or instance, and sparse coding is a powerful tool for
feature/image reconstruction\cite{Ref:Yang2009Linear,Ref:feng2017multi} and similarity measurement~\cite{Ref:yan2009semi,Ref:Yang2014Data}.
Besides, it has also been successfully applied in text mining~\cite{Ref:li2015reader}, speech signal processing~\cite{Ref:Zhen2016Under}, recommendation~\cite{Ref:Feng2016Social}, etc.

Basically, given a data matrix $X = [{x}_{1}, {x}_{2}, \dots, {x}_{n} ] \in \mathbb{R}^{m \times n}$,
consisting of $n$ data points measured on $m$ dimensions, $x_{i}=[ x_{i}^{1}, x_{i}^{2}, \dots, x_{i}^{m}]^{T} \in \mathbb{R}^{m}$ is a vector representing the $i$-th data point.
Sparse Coding (SC)~\cite{Ref:SC} aims at extracting a dictionary (also called a base) that consists of a set of atom items (called a basis), and converts data instances into sparse coordinates with respect to the dictionary.
In other words, it decomposes the data matrix $X$ into a dictionary matrix $B = [{b}_{1}, {b}_{2}, \dots, {b}_{r}] \in \mathbb{R}^{m \times r}$
and a sparse code matrix $S = [{s}_{1}, {s}_{2}, \dots, {s}_{n}] \in \mathbb{R}^{r \times n}$,
so that each data sample ${x}_{i}$ can be represented as a sparse linear combination of atom items
(i.e., ${b}_{i}$) in the dictionary $B$ with corresponding sparse coefficient vector ${s}_{i}$.
Formally, the sparse coding problem can be described as the following optimization problem:
\begin{eqnarray}\label{sc}
\min \limits_{B, S} && O^{SC} = \left\| {X - BS} \right\|_F^2 + \beta \sum_{i = 1}^n \|{s}_i\|_1 \\
\textrm{s.t.} && \left\| {b}_j \right\|_2^2 \leq 1, \forall j = 1,2..,r \nonumber
\end{eqnarray}
where $\| \cdot \|_F$ denotes the matrix Frobenius norm (i.e., $\left\| A \right\|_F = \sqrt{ \sum_i \sum_j a_{ij}^2}$, $\forall A$),
$\| \cdot \|_1$ denotes the $\ell_1$ norm of a given vector (i.e., $\left\| {a} \right\|_1 =  \sum_{i=1}^n | a_{i} |$, $\forall {a}$),
and $\beta$ is the sparsity regularization parameter that represents the tradeoff between reconstruction error and sparsity.

Naturally, the optimization of sparse coding in (\ref{sc}) is not jointly convex for both $B$ and $S$,
but it is convex in $B$ (while holding $S$ fixed) and convex in $S$ (while holding $B$ fixed) separately.
Thus, it is usually solved iteratively by alternatingly minimizing over $B$ and $S$, while holding the other fixed.
In this way, when $B$ is fixed, (\ref{sc}) becomes an $\ell_1$-regularized non-smooth convex optimization problem and many efforts have been done for solving this type of problem~\cite{Ref:yang2010fast}.
When $S$ is fixed, it is a well-defined constrained least squares problem, which can be solved efficiently by existing optimization methods~\cite{Ref:SC}.

Though we can optimize SC by an alternative method, there is a big limitation caused by the non-convexity of the objective in (\ref{sc}).
This deficiency will drive the current SC optimization approach get stuck into bad local minima, especially when many noises and outliers exist in the datasets.
To relieve the affect of non-convexity, a heuristic approach is to run the algorithm multiple times with different initializations or learning rates, and pick the best results based on cross-validation. However, this strategy is ad hoc and generally inconvenient to implement in unsupervised setting, since there is
no straightforward criterion for choosing a proper solution.
Thus, for robust SC optimization, our goal is to alleviate the influence of the noisy and confusing data, as the confusing data usually correspond to the highly nonlinear local patterns which is hardly learnable for the model space, and the noisy ones are the outliers that should not be learned, such as noisy pixels in images, or noisy words in texts.
Typically, this learning robustness will be achieved by a well sample selection to distinguish the reliable samples from the confusing ones.
Fortunately, a recently studied learning regime, named $Self$-$Paced~Learning$ (SPL)~\cite{Ref:Kumar2010Self} is making effort for this issue.
The core idea of SPL is to learn the model with easier samples first and then gradually process more complex ones, which is simulating the learning principle of humans/animals.
This learning mechanism is empirically demonstrated to be beneficial for some learning tasks~\cite{Ref:xu2015multi,Ref:Zhao2015Self,Ref:li2017self}, where it can in avoid bad local minima and achieve a better generalization result.
Therefore, incorporating it into SC is expected to alleviate the local minimum issue of alternative learning.

To enhance the learning robustness, in this paper, we propose a unified framework named $Self$-$Paced~Sparse~Coding$ (SPSC).
With an adaptive pace from easy to hard in optimization, SPSC can dynamically introduce the data values into learning from easy ones to complex
ones.
Specifically, for different levels of dynamic selection, we extends SPSC into three variants, that is $Sample$-$wise$ SPSC, $Feature$-$wise$ SPSC and $Element$-$wise$ SPSC.
We also present a simple yet effective algorithm framework for solving the proposed SPSC problems.
Experimental results on the real-world clean datasets and those with corruption demonstrate the effectiveness of the proposed approach, compared to the state-of-the-arts.

\section{Related Work}
In this section, we will review related works on sparse coding and self-paced learning, as our work $Self$-$Paced~Sparse~Coding$ lies in cross road of these two fields.
\subsection{Sparse Coding}
Sparse coding consists of two components, that is, the optimization of the corresponding sparse codes (as dictionary $B$ is known) and the learning of optimal dictionary (as sparse code $S$ is given).
For sparse code optimization, when the sparsity of codes is measured as $\ell_0$-norm, it can be solved by methods such as matching pursuit, orthogonal matching pursuit, basis pursuit, etc; and multiple approaches are presented for $\ell_1$-norm ionization, such as coordinate descent~\cite{wu2008coordinate},
interior-point method~\cite{kim2007interior}, active set method~\cite{lee2007efficient}, and augmented lagrange multiplier method~\cite{lin2010augmented}

The dictionary is usually automatically learned from data rather than by a predetermined dictionary.
Typically, sparse coding is suitable to reconstruct data, and consequently variants of traditional sparse coding are presented for different purposes:
(i) Structure regularized sparse coding.
Structured sparse coding methods exploit different structure priors among the data.
\cite{Ref:Sun2014Structured} incorporate additional structured sparsity to the sparse coding objective functions, which leads to the promising performances on the applications.
Graph/Hypergrpah regularized sparse coding \cite{Ref:GSC,Ref:HLSC,Ref:feng2017multi} are proposed to preserve the intrinsic manifold
structure among high dimensional data objects, and impose Laplacian regularization as a manifold regularization on sparse coefficients.
(ii) Supervised/Semi-supervised sparse coding. In this group, label information of a training dataset is leveraged to learn a discriminative dictionary for better classification performances.
Approaches in \cite{Ref:FisherDL} attempt to learn multiple dictionaries or category-specific dictionaries to promote discrimination between classes.
Others in \cite{Ref:SubDL2} learn a dictionary shared by all classes by merging or selecting atom items from an
initially large dictionary or incorporates discriminative terms into the objective function during training~\cite{Ref:DKSVD,Ref:LCKSVD,Ref:FDKSVD}.
(iii) Muli-modal sparse coding. In this group, traditional sparse coding approaches can be extended to benefit from the information encoded in multiple
sources for cross-domain sparse representation~\cite{Ref:Jing2014Uncorrelated,Ref:Li2017Multi}.

Most of the current SC models are developed on nonconvex optimization problems, and thus always encounter the local minimum problem, especially when there are noisy and outlier data. We thus aim to alleviate this issue by advancing it into the SPL framework.
\subsection{Self-Paced Learning}
Inspired by the intrinsic cognitive principle of humans/animals, Bengio et al.~\shortcite{Ref:bengio2009curriculum} initialized a new learning paradigm called $curriculum learning$ (CL), the core idea of which is to learn a model by gradually including samples into training from easy to complex for training non-convex objectives.
This learning process is beneficial for alleviating local solation  in non-convex optimization, as supported by empirical evaluation~\cite{Ref:basu2013teaching}.
However, it requires a priori identification to identify the easy and the hard samples in advance, which is difficult in real-world applications.
To overcome the limitation due to the heuristic sample easiness ranking as in CL, Kumar et al.~\shortcite{Ref:Kumar2010Self} proposed a concise model, called $Self$-$Paced~Learning$ (SPL) to introduce a regularizor into the learning objective.
SPL can automatically optimize an appropriate curriculum by the model itself with a weighted loss term on
all samples and a general SPL regularizer imposed on sample weights.
Jiang et al.~\shortcite{Ref:jiang2015self} presented a more comprehensive model learning underlying SPL/CL, as a general optimization problem as
\begin{eqnarray}\label{spl}
\min \limits_{w, v} &&\mathbb{E}(w,v,\lambda) = \sum_{i=0}^{n} {v_i \ell(y_i;g(x_i,w))} + f(v;\lambda)  \nonumber \\
\textrm{s.t.} && v \in \Psi
\end{eqnarray}
where a training set $\mathcal{D} = \{(x_i,y_i)\}_{i=1}^n$; $\ell(y_i;g(x_i,w))$ denotes the loss function which calculates the cost between the true label $y_i$ and estimated label $g(x_i,w)$ under a model $g$ with $w$ as its model parameter to learn; $v = [v_1, v_2,\ldots, v_n]^T \in [0,1]^n$ denotes the weight variables reflecting the samples¡¯ confidence; $f$ corresponds to a self-paced regularizer function for the sample selection scheme; $\Psi$ is a feasible region that encodes the information of a predetermined curriculum; $\lambda$ is a parameter to control the learning pace, also referred as "pace age". For the self-paced regularizer function $f(v;\lambda)$, Jiang et al.~\shortcite{Ref:jiang2015self} abstracted three necessary condition should be satisfied.

By sequentially optimizing the model with gradually increasing pace parameter on the SPL regularizer, more samples can be automatically discovered in a pure self-paced way. The number of samples selected at each iteration is determined by a weight that is gradually annealed such that later iterations introduce more samples. The algorithm converges when all samples have been considered and the objective function cannot be improved further.
By virtue of its generality, SPL have been considered in a broad spectrum of latent variable models, such as matrix factorization~\cite{Ref:Zhao2015Self}, clustering~\cite{Ref:xu2015multi}, multi-task learning~\cite{Ref:li2017self}, dictionary learning~\cite{Ref:tang2012self,Ref:xu2016multi}.
Opposite to existed self-paced dictionary learning, in this paper we try to utilize the generalized SPL sheila and implement a soft sample selection accordingly rather than a heuristic hard strategy sample selection in~\cite{Ref:tang2012self}.
Also, in~\cite{Ref:xu2016multi}, it only address the issue of bad local minimum of non-convex sparse coding optimization by assigning weights to different samples of training data.
However, due to the complexity and noisy nature of real data, the confidence levels of features of the dataset may also vary (eg., features extracted with different approaches for a set of images should have different confidence level), and the confidence levels of the specific values in the same sample also vary (eg., in image denoising task, noisy pixels should have lower confidence level comparing to the uncorrupted ones in the same image).
Thus, considering a unified self-paced learning for sparse coding may be the further direction.

\section{Self-Paced Sparse Coding}
\subsection{The Formulation of SPSC}
Generally, the objective of SC is to learn a dictionary $B$ and representation $S$ from $X$ such that it can well reconstruct the data and $S$ can follow some prior knowledge as in (\ref{sc}).
Inspired by the fact that humans often learn concepts from the easiest to the hardest as in SL and SPL, we incorporate the easy-to-hard strategy operated on the samples into the learning process of SC, as $Self$-$Paced~Sparse~Coding$ (SPSC).
Thus, following the framework of SPL in (\ref{spl}), the new learning objective function of SPSC can be formulated as:
\begin{eqnarray}
\label{objetiveSPSC}
\min \limits_{B,S,v} &&\|(X-BS) \sqrt{\mathbf{Diag}(v)}\|_F^2 + f(v;\lambda) +  R(S) + \beta\|S\|_1 \nonumber \\
\textrm{s.t.} && \left\| {b}_j \right\|_2^2 \leq 1, \forall j = 1,2..,r; v^{i} \in [0,1].
\end{eqnarray}
where $v = [v_1, v_2,\ldots, v_n]^T \in [0,1]^n$ denotes the weight vector imposed
on samples in $X$; let '$\sqrt{\cdot}$' denote element-wise square root of a matrix;
`$\mathbf{Diag}(\cdot)$' denotes the diagonal matrix with diagonal entries as those in the vector, that is,
\[
\|(X-BS) \sqrt{\mathbf{Diag}(v)} \|_F^2 = \sum_{i=0}^{n} v_{i} \|x_{i}-Bs_i\|_2^2.
\]
It is noted that, different from the original SPL learning framework as in (\ref{spl}), as sparse coding is an originally unsupervised learning, the loss function $\ell$ corresponds to $\ell_{i} = \|x_{i}^{j}-Bs_i\|_2^2$, which measures the reconstruction error between $x_i$ and estimated $Bs_i$ after learning.
\subsection{SPL Regularizer}
For regularizer $f(v;\lambda)$, the original SPL~\cite{Ref:Kumar2010Self} adopted negative $\ell_1$-norm as
\[
f(v;\lambda) = -\lambda \|v\|_1 = -\lambda \sum_{i=0}^{n}v_{i}.
\]
Under this regularizer, with fixed $(B,S)$, the global optimal $v_{i}$ is calculated by the optimization as
\begin{equation}\label{slpR}
\min \limits_{v_{i} \in [0,1]}  \sum_{i=0}^{n} v_{i}\ell_{i} -f (v;\lambda).
\end{equation}
Then, the solution can be written as a hard threshold schema:
\begin{equation}
\label{slpSolutionHard}
v_{i} = \left \{ {\begin{array}{l}
1, ~~ \ell_{i}<\lambda \\
0, ~~ \ell_{i}\geq\lambda\\
 \end{array}} \right.
\end{equation}
Instead of hard weighting, in our experiments, we utilize the linear soft weighting regularizer~\cite{Ref:jiang2015self} due to its relatively easy implementation and well adaptability to complex scenarios.
This regularizer penalizes the sample weights linearly in terms of the loss. Specifically, we have
\begin{equation}
\label{slpSoft}
f(v;\lambda) =\lambda (\frac{1}{2}\|v\|_2^2 - \|v\|_1) = \sum_{i=1}^{n}(\frac{1}{2}v_{i}^2-v_{i}).
\end{equation}
(\ref{slpSoft}) is convex with respect to $v_{i}$, and thus the global minimum can be obtained with analytical solution for the linear soft
weighting, as,
\begin{equation}
\label{slpSolutionSoft}
v_{i} = \left\{
 {\begin{array}{l}
1-\frac{\ell_{i}}{\lambda},~\ell_{i}<\lambda\\
 0,~~\ell_{i} \geq \lambda\\
 \end{array}}
\right.
\end{equation}
It is seen that if an sample can be well reconstructed ($\ell_{i}<\lambda$), it will be selected as an easy sample ($v_{ij}>0$) and the weight is decreasing with respect to the reconstruction error, or otherwise unselected ($v_{i}=0$).
The parameter $\lambda$ controls the pace at which the model learns new samples, and physically it can be interpreted as the "age" of the model.
When $\lambda$ is smaller, more samples will be unselected as their corresponding losses are more likely to be greater than $\lambda$.
As $\lambda$ becomes larger, more samples with larger losses will be gradually considered.
\subsection{Regularization on Sparse Codes}
Prior relational knowledge among the samples is very useful when learning new representations.
To embed this knowledge into sparse coding, we will adopt the newly proposed manifold regularizer, called hypergraph
consistency regularization (HC)~\cite{Ref:feng2017multi}, which is empirically shown to perform better in sparse coding.
Specifically, it tries to reconstruct the hyperedge incidence matrix using sparse codes $S$ in addition to the Laplcaican regulation on $S$ as
\begin{equation}
\label{regulS}
R(S) = \gamma \| I - QS\|_F^2 + \alpha tr(SLS^T),
\end{equation}
where, $I\in \{0,1\}^{|E|\times |V|}$ is the incidence matrix of hypergraph $G(X, E)$ which consists of a set of vertices $X = [{x}_1, {x}_2, \dots, {x}_{n}]$, and a set of hyperedges $E = [{e}_1, {e}_2, \dots, {e}_{|E|}]$.
The hypergraph can be constructed using the original data of $X$ to record locally geometrical structure of the original data; $I_{ji}= 1$, if ${x}_i \in {e}_j$; $I_{ji}= 0$, otherwise. More generally, $I$ can also be defined in a probabilistic way~\cite{Ref:PhyperGrpah}, such that $I_{ij} \in [0,1]$.
$L$ denotes the Laplacian matrix defined as $L = \mathbb{I} - W$ ($\mathbb{I}$ is the identity matrix); $W \in \mathbb{R}^{n \times n}$ is the weight matrix of a hyergraph to measure how close each pair of two vertices in the manifold space are, which can be defined as follows (with normalization):
$W = D_x^{ - 1 / 2} I^T W_e D_e ^{ - 1} I D_x^{ - 1 / 2},$
where $D_{e}$ and $D_{x}$ respectively denote the diagonal matrices of hyperedge degrees and vertex degrees,
and $W_{e}$ denotes a diagonal matrix of the hyperedge weights.
By minimizing $R(S)$, the incidence matrix $I$ can be well reconstructed and the consistency of the hypergraph structure on sparse code space $S$ is guaranteed.
\subsection{SPL Selection on Different Levels}
It is seen that SPSC in (\ref{objetiveSPSC}) is implementing sample-level selection during the learning process, where the sample can be an image instance, a text, or a user in various tasks. Accordingly, we note the SPSC in (\ref{objetiveSPSC}) as $Sample$-$wise$ SPSC (SPSC$^S$).
Usually, a sample $x$ is represented by a multi-dimensional vector $x=[x^1,x^2,\ldots,x^m]\in \mathbb{R}^n$ with each element $x^j$ as a partial description of $x$. As in in (\ref{objetiveSPSC}), each element is treated the same, that is, equal confidence for model learning.
However, due to the complexity and noisy nature of real data, the confidence levels of the specific element in the same sample also vary (eg., in image denoising task, noisy pixels should have lower confidence level comparing to the uncorrupted ones in the same image).
Also the confidence levels of features could be also different (eg., features extracted with different approaches for a set of images should corresponds to different confidence level),
We then show how to incorporate the easy-to-hard strategy operated on elements and features into the learning process of SC.
\subsubsection{$Element$-$wise$ SPSC} When operating SPSC on element-level learning, the loss for each element $x_i^j$ and could be defined as $\ell_{ij} = (x_i^j-b^js_i)^2$. Then the learning objective of $Element$-$wise$ SPSC (SPSC$^E$) becomes the following formulation:
\begin{eqnarray}
\label{objetiveSPSC_Ele}
\min \limits_{B,S,V} &&\|(X-BS)\odot \sqrt{V}\|_F^2 + f(V;\lambda) +  R(S) + \beta\|S\|_1 \nonumber \\
\textrm{s.t.} && \left\| {b}_j \right\|_2^2 \leq 1, \forall j = 1,2..,r; v_{ij} \in [0,1].
\end{eqnarray}
Where $V$ represents the matrix composing of $v_{ij}$ which denotes the weights imposed on the observed elements of $X$; `$\odot$' denotes element-wise multiplication, that is,
\[
\|(X-BS)\odot \sqrt{V}\|_F^2 = \sum_{i=0}^{n}\sum_{j=0}^{m} v_{ij}(x_{i}^{j}-b^{j}s_i)^2.
\]
Here $b^j$ denotes the $j$-th row vector of dictionary $B$. Thus, $\|(X-BS)\odot V\|_F^2$ is the self-paced regularizer determining the elements to be selected in learning. To defines $f(V;\lambda)$, the SPL regularizer in (\ref{slpSoft}) can be easily generalized as
\begin{equation}
\label{slpSoft_Eel}
f(V;\lambda) =\lambda (\frac{1}{2}\|V\|_F^2 - \|V\|_1) = \sum_{i=1}^{n}\sum_{j=1}^{m}(\frac{1}{2}v_{ij}^2-v_{ij}).
\end{equation}
Accordingly, the optimal solution of $V$ can be obtained by
\begin{equation}
\label{slpSolutionSoft_Ele}
v_{ij} = \left\{
 {\begin{array}{l}
1-\frac{\ell_{ij}}{\lambda},~\ell_{ij}<\lambda\\
 0,~~\ell_{ij} \geq \lambda\\
 \end{array}}
\right.
\end{equation}
\subsubsection{$Feature$-$wise$ SPSC} When operating SPSC on feature-level learning, the learning objective of $Feature$-$wise$ SPSC (SPSC$^F$) is as the following optimization:
\begin{eqnarray}
\label{objetiveSPSC_Fea}
\min \limits_{B,S,v} &&\|\sqrt{\mathbf{Diag}(v)}(X-BS)\|_F^2 + f(v;\lambda) +  R(S) + \beta\|S\|_1 \nonumber \\
\textrm{s.t.} && \left\| {b}_j \right\|_2^2 \leq 1, \forall j = 1,2..,r; v_{i} \in [0,1].
\end{eqnarray}
where $v = [v^1, v^2,\ldots, v^n]^T \in [0,1]^n$ denotes the weight vector imposed
on different features in $X$; the loss of each feature corresponds to $\ell^{i} = \|x^{j}-b^jS\|_2^2$, which measures the reconstruction error of each feature and $x^j$ represents the $j$-th row vector in $X$, that is, a feature value across all samples.
The SPL regularizer and optimal weight canaliculation in (\ref{slpSoft}) and~(\ref{slpSolutionSoft}) can be easily adopted to SPSC$^F$, as the formulations in (\ref{objetiveSPSC_Fea}) and (\ref{objetiveSPSC}) are mathematically equivalent to each other.
\section{Optimization of SPSC}
\subsection{Optimization Algorithm}
Comparing the proposed three formulations of SPSC in (\ref{objetiveSPSC}),~(\ref{objetiveSPSC_Ele}) and~(\ref{objetiveSPSC_Fea}), SPSC$^E$ in  (\ref{objetiveSPSC_Ele}) can be seen as a general form of these three.
Thus, in the following we will show how to optimize (\ref{objetiveSPSC_Ele}).
The optimization problem is not jointly convex, so we will use alternative search strategy (ASS) to solve it, suggested in~\cite{Ref:Kumar2010Self}.
Following ASS, we can alternatively optimizes $V, B, S$ while keeping the other set of variables fixed.

(i) Solve for $V$: With fixed $B$ and $S$, optimization of selection weight matrix $V$ can be solved under (\ref{slpSolutionSoft_Ele}).

(ii) Solve for $B$: With fixed codes $S$ and weight matrix $V$, the optimization problem for dictionary $B$ writes:
\begin{equation}
\label{optimizeB}
\min \limits_{B}\|(X-BS)\odot \sqrt{V}\|_F^2 ~~s.t.\left\| {b}_j \right\|_2^2 \leq 1, \forall j = 1,2..,r.
\end{equation}
This problem is a Quadratically Constrained Quadratic Program (QCQP) which can be solved using Lagrangian dual~\cite{Ref:SC}.

(iii) Solve for $S$: With fixed dictionary $B$ and weight matrix $V$, combing (\ref{objetiveSPSC_Ele}) and (\ref{regulS}), the optimization problem for codes $S$ corresponds to the weighted SC:
\begin{equation}
\label{optimizeS}
\min \limits_{S}\|(X-BS)\odot \sqrt{V}\|_F^2 +  \gamma \| I - QS\|_F^2 + \alpha tr(SLS^T) + \beta \|S\|_1.
\end{equation}
This problem is a $\ell_1$-regularized convex optimization and off-the-shelf algorithms can be employed for solving it, such as feature-sign algorithm in~\cite{Ref:HLSC}.
The whole process is summarized in Algorithm 1.
\begin{table}[!h]
\SetKwInOut{Input}{Input}
\SetKwInOut{Parameter}{Param}
\footnotesize
\centering
\begin{tabular}
{p{210pt}}
\hline
$\mathbf{Algorithm~1}$ Self-Paced Sparse Coding (SPSC)\\
\hline
\Input{Data matrix $X \in \mathbb{R}^{ m \times n}$ and a hypergraph $G$ with $I$ and $L$ as its corresponding incidence matrix and weight matrix;
nonnegative trade-off parameters $\alpha, \beta, \gamma$ and stepsize $\mu > 1$.}\;
1: Initializing: solve the SC problem with all elements selected ($v_{ij} = 1$) to obtain optimal $B^{(0)}$ and $S^{(0)}$;
calculate the loss $\ell_{ij}$ of all elements; $t \leftarrow 0$\; initialize self-paced parameters $\lambda$;

2: $\mathbf{repeat:}$\;

3: \hspace{0.5cm} Update $V^{(t+1)}$ via (\ref{slpSolutionSoft_Ele}).\;

4: \hspace{0.5cm} Update $B^{(t+1)}$ via Solving (\ref{optimizeB}).\;

5: \hspace{0.5cm} Update $S^{(t+1)}$ via Solving (\ref{optimizeS}).\;

6: \hspace{0.5cm} Compute current loss $\ell_{ij}$.\;

7: \hspace{0.5cm} $\lambda \leftarrow \lambda \mu $, $t \leftarrow t+1$. \tcc{updating the learning pace}\;

8: $\mathbf{until}$ convergence. \;

9: \Return{$B = B^{(t)}$,$S = S^{(t)}$} \;\\
\hline
\end{tabular}
\label{tab:algorithm}
\end{table}
The algorithm converges when all samples have been considered and the objective function cannot be improved further.
It is seen from the algorithm that the number of samples selected at each iteration is determined by a weight that will be
gradually annealed such that later iterations introduce more elements.
For optimization of SPSC$^S$ or SPSC$^F$, in Step 6, it can be easily solved by respectively calculating $sample$-$wise$ loss $\ell_i$ or $Feature$-$wise$ loss $\ell^j$ defined above. For weight update in in Step 3, all weights of elements are set equal to the weights of corresponding sample or feature, i.e. $v_{ij} = v_i$ or $v_{ij} = v^j$.
%\subsection{Convergence and Computational Complexity}
%We first analyze the convergence property of Algorithm 1.
During each iteration, an convex minimization problem is solved for each V-step, B-step and S-step, and the global
optimum solution is obtained for each sub-step.
Thus the overall objective is non-increasing with the iterations and Algorithm 1 is guaranteed to convergence.
%See- Self-Paced Multi-Task Learning; Robust Localized Multi-view Subspace Clustering

\section{Experiment Results}
In this section, we extensively evaluate the proposed approach when applied to image clustering task on two real datasets.
Experimental results demonstrate the correctness and effectiveness of the proposed model.
\subsection{Experiment Setup}
\subsubsection{Datasets}
Two widely used real-world image dataset benchmarks are considered in the experiments\footnote{We downloaded these datasets from:\url{http://www.cad.zju.edu.cn/home/dengcai/Data}.}.
\begin{itemize}
  \item {\bf COIL20}.
      The COIL20 image library contains $1,440$ images of $20$ objects,
      with $72$ images per object.
      The size of each image is $32 \times 32$, with $256$ grey levels per pixel.
      Images of the objects were taken at angle intervals of $5^{\circ}$.
      Each image is represented by a $1024$-dimensional vector.
  \item {\bf CMU-PIE}.
      The CMU-PIE face database contains $41,368$ facial images of $68$ subjects in various angles, facial expressions, and lighting conditions.
      The size of each image is $32\times 32$, with $256$ grey levels per pixel.
      Thus, each image is represented by a $1024$-dimensional vector.
      In our experiments, we fix the angle and the facial expression to select 21 images under different lighting conditions for each subject,
      totaling $1,428$ images.
\end{itemize}
\subsubsection{Baselines}
We compared the proposed approach with several state-of-the-art methods.
\begin{itemize}
  \item {\bf Original} and {\bf SC}.
      The ``original'' method is to cluster image objects in the original data space (denoted as 'OS'). Sparse coding  \cite{Ref:SC} is the basic sparse coding method without any regularization except sparsity constraints.
  \item {\bf LSC$^G$}, {\bf LSC$^H$} and {\bf CSC}.
      (LSC$^G$) \cite{Ref:GSC} and (LSC$^H$) \cite{Ref:HLSC} add a graph/hyperGraph Laplacian regularization term to the original sparse coding framework,
      while CSC~\cite{Ref:feng2017multi} (hypergraph consistent sparse coding) integrates a hypergraph incidence consistency regularization term.
  \item {\bf SPSC$^{S}$}, {\bf SPSC$^{E}$} and {\bf SPSC$^F$}.
  These are proposed methods as in~(\ref{objetiveSPSC}),~(\ref{objetiveSPSC_Ele}) and~(\ref{objetiveSPSC_Fea}) respectively.
\end{itemize}
\subsubsection{Experimental Setting}
We use $k$-means clustering to group image datasets and compare the clustering results with two commonly used metrics, i.e., clustering accuracy (ACC)
and normalized mutual information (NMI).
Both ACC and NMI range from 0 to 1, while ACC reveals the clustering accuracy (also called purity) and
 NMI indicates whether the different clustering sets are identical (NMI = 1) or independent (NMI = 0).
To ensure stability of results, the direct $k$-means clustering on the sparse code space is respectively repeated 30 times, each time with a random set of initial cluster centers.
The average ACC and NMI for different methods over these 30 repetitions on each dataset will be reported.
\begin{table*}
\caption{Clustering performance on COIL20 dataset using different methods}
\scriptsize
\centering
\label{Table:resultCOIL}{
\begin{tabular}
{c|c c|c c|c c|c c|c c|c c|c c}
\hline
\raisebox{-1.50ex}[0cm][0cm]{$\rho$}&
\multicolumn{2}{|c|}{OS} &
\multicolumn{2}{|c|}{SC} &
\multicolumn{2}{|c|}{HLSC} &
\multicolumn{2}{|c|}{HIC} &
\multicolumn{2}{|c|}{SPSC$^E$} &
\multicolumn{2}{|c|}{SPSC$^S$} &
\multicolumn{2}{|c}{SPSC$^F$}  \\
\cline{2-15}
 &ACC&NMI&ACC&NMI&ACC&NMI&ACC&NMI&ACC&NMI&ACC&NMI&ACC&NMI \\
\hline
0&
0.6120 &
0.7353 &
0.6022 &
0.7291 &
0.6967 &
0.8035 &
0.7320 &
0.8729 &
0.7828 &
0.8999 &
0.7889 &
0.9043 &
0.7863 &
0.8973  \\
\hline
0.2&
0.6081 &
0.7323 &
0.5847 &
0.7151 &
0.6742 &
0.8025 &
0.7319 &
0.8668 &
0.7878 &
0.8953 &
0.7580 &
0.8989 &
0.7726 &
0.8895  \\
\hline
0.4&
0.6043 &
0.7303 &
0.5807 &
0.7172 &
0.6888 &
0.8068 &
0.7258 &
0.8820 &
0.7737 &
0.8920 &
0.7576 &
0.8886 &
0.7511 &
0.8826  \\
\hline
0.6&
0.5968 &
0.7294 &
0.5688 &
0.7187 &
0.6806 &
0.8068 &
0.7237 &
0.8726 &
0.7655 &
0.8963 &
0.7545 &
0.8843 &
0.7563 &
0.8826  \\
\hline
0.8&
0.5819 &
0.7132 &
0.5868 &
0.7115 &
0.6705 &
0.7927 &
0.7122 &
0.8540 &
0.7636 &
0.8851 &
0.7474 &
0.8665 &
0.7484 &
0.8740  \\
\hline
1&
0.5879 &
0.7103 &
0.5794 &
0.7151 &
0.6790 &
0.7903 &
0.7042 &
0.8493 &
0.7624 &
0.8807 &
0.7433 &
0.8737 &
0.7449 &
0.8752  \\
\hline
\end{tabular}}
\end{table*}
\begin{table*}
\caption{Clustering performance on MNIST dataset using different methods}
\scriptsize
\centering
\label{Table:resultMNIST}{
\begin{tabular}
{c|c c|c c|c c|c c|c c|c c|c c}
\hline
\raisebox{-1.50ex}[0cm][0cm]{$\rho$}&
\multicolumn{2}{|c|}{OS} &
\multicolumn{2}{|c|}{SC} &
\multicolumn{2}{|c|}{HLSC} &
\multicolumn{2}{|c|}{HIC} &
\multicolumn{2}{|c|}{SPSC$^E$} &
\multicolumn{2}{|c|}{SPSC$^E$} &
\multicolumn{2}{|c}{SPSC$^E$}  \\
\cline{2-15}
 &
ACC&
NMI&
ACC&
NMI&
ACC&
NMI&
ACC&
NMI&
ACC&
NMI&
ACC&
NMI&
ACC&
NMI \\
\hline
0&
0.5656 &
0.5252 &
0.5076 &
0.5074 &
0.6643 &
0.6411 &
0.6678 &
0.6679 &
0.6814 &
0.6871 &
0.6816 &
0.6791 &
0.6887 &
0.6792  \\
\hline
0.2&
0.5649 &
0.5234 &
0.5252 &
0.5056 &
0.6551 &
0.6452 &
0.6670 &
0.6611 &
0.6898 &
0.6823 &
0.6765 &
0.6675 &
0.6785 &
0.6820  \\
\hline
0.4&
0.5555 &
0.5192 &
0.5298 &
0.5054 &
0.6554 &
0.6393 &
0.6613 &
0.6673 &
0.6762 &
0.6825 &
0.6728 &
0.6527 &
0.6689 &
0.6767  \\
\hline
0.6&
0.5426 &
0.5084 &
0.5189 &
0.4932 &
0.6420 &
0.6212 &
0.6423 &
0.6363 &
0.6659 &
0.6717 &
0.6550 &
0.6538 &
0.6546 &
0.6615  \\
\hline
0.8&
0.5442 &
0.5021 &
0.5175 &
0.4940 &
0.6381 &
0.6171 &
0.6440 &
0.6314 &
0.6682 &
0.6604 &
0.6512 &
0.6471 &
0.6557 &
0.6539  \\
\hline
1&
0.5362 &
0.4939 &
0.5128 &
0.4888 &
0.6239 &
0.6130 &
0.6362 &
0.6326 &
0.6509 &
0.6580 &
0.6382 &
0.6477 &
0.6503 &
0.6511  \\
\hline
\end{tabular}}
\end{table*}
For each dataset, we normalize each image vector into a unit $\ell_2$-norm as the input of comared sparse coding algorithms.
The dictionary size $r$ of all these models is set to be $128$,
since several recent works on sparse coding have advocated the use of overcomplete representations for images \cite{Ref:olshausen1996emergence,Ref:GSC}.
Thus, after sparse coding, each image is represented as a $128$-dimension vector,
and will be used as the input of $k$-means clustering ($k$-means).
% and normalized cuts based spectral clustering (Ncuts)
%Similar to \cite{Ref:scsparse}, we use cosine similarity to compute the weight between images for spectral clustering.
The graph in LSC$^G$ and hypergraph in CSC and SPSC are constructed in the original feature space after unit-norm normalization.
Specifically, The $k$-nearest neighbor-based graph (or hypergraph) (with $k=3$ by Euclidean distance) is used to characterize the intrinsic manifold with a binary weighting scheme as suggested in \cite{Ref:GSC,Ref:HLSC}.

The parameters are tuned as follows.
For SC, the sparsity weight $\beta$ is tuned in the range of $[0.005,0.01,0.02,0.04,0.08]$ to get the best performance (i.e., the highest average of ACC in 30 runs),
and the best value is used for all sparse coding framework-based methods.
For LSC, the Laplacian regularization weight $\alpha$ is tuned in the range of $[0.5,1,2,4,8,16,32,64]$.
For CSC, we respectively fix the best values of $\alpha$ and $\beta$, and tune $\gamma$.
For SPSC, we use the best $\alpha$, $\beta$ and $\gamma$ of CSC; and we initialize $\lambda$ to a value when $50\%$ of the elements (samples, features) are selected in the first iteration and set stepsize $\mu =1.2$.

To test the robustness of proposed SPSC to corruption using COIL20 dataset.
For each image $x$ with pixel value representation, we add white Gaussian noise according to $\widetilde{x} = x + \rho \mathbf{n}$,
where $\mathbf{n}$ is the noise following a normal distribution with zero average and $\sigma$ is set to be the standard deviation over entries in data matrix $X$,
$\rho$ is a corruption ratio which increase from $0.2$ to $1.0$ in intervals of $0.2$. Clearly, white Gaussian noise is additive.
Figure.~\ref{fig:figNoise} shows some samples on COIL20 dataset.
\begin{figure}[h]
\centering
\subfigure {\epsfig{file=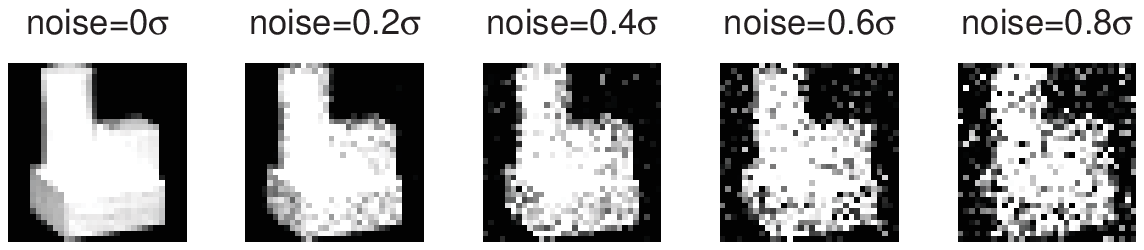,width=0.48\textwidth}}
\subfigure {\epsfig{file=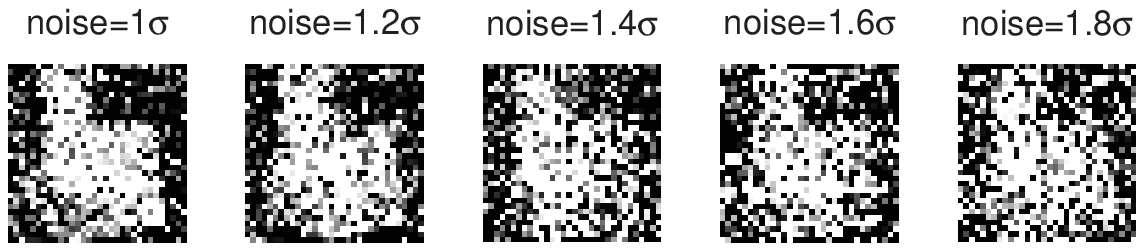,width=0.48\textwidth}}
\caption{COIL20 images with corruptions by Gaussian noise.}
\label{fig:figNoise}
\end{figure}
\subsection{Results and Discussion}
Tables~\ref{Table:resultCOIL} and~\ref{Table:resultMNIST} report the performance of compared methods on clustering. They show the following points.
1) All three variants of proposed SPSC perform better than other baselines on both ACC and NIM, the power of self-paced learning is demonstrated.
2) Among three variants of proposed SPSC, $element$-$wise$ SPSC (SPSC$^S$) is more robust than $sample$-$wise$ SPSC and $sample$-$wise$, especially when the noise level is larger. We think that this is because the $element$-$wise$ SPSC is implementing on the dynamic selection on the smallest granularity.
3) When the noise is on a low level ($\rho < 1$), the clustering performance will be not severely affected.
 and however when the noise level is high ($\rho > 1$), the clustering performance will sharply decrease (we implement experiments for $\rho > 1$ but don't show it due to the lower accuracy).
\subsection{Discussion on the Weight Learning}
In the experiment, we record the weight matrix $V$ in each iteration in Algorithm 1. To show the process of dynamic selection of samples, we plot the correlation between noise level (the mean absolute value of noise over each sample) and the corresponding weight in each iteration, as in Figure.~\ref{fig:figNoise}.
\begin{figure}[h]
\epsfig{file=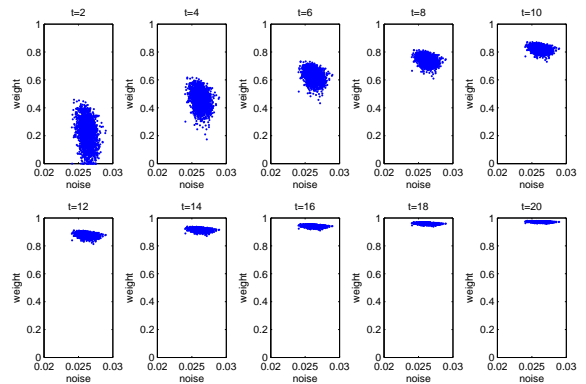,width=0.52\textwidth}
\centering
\caption{The correlation between noise level and weight in each run.}
\label{fig:figNoise}
\end{figure}
We can see that, the values in the weight matrix are becoming larger after each iteration and they keep a high value after about 15 iterations, following the mechanism of self-paced learning that more elements are selected after each iteration of the optimization process.
Also, there is a negative correlation between noise level and the corresponding weight in the beginning of iterations, means the algorithm of SPSC is learning samples with lower noise corruption (easier samples) first.

\section{Conclusion}
Under the framework of alternative optimization of sparse coding, the non-convexity usually drive the optimization approach get stuck into bad local minima, especially when many noises and outliers exist in the datasets. 
To relieve the affect of non-convexity, we propose a unified framework named $Self$-$Paced~Sparse~Coding$ (SPSC) by incorporating Self-paced learning methodology with Sparse coding as well as manifold regularization.
For different levels of dynamic selection, we extends SPSC into three variants, that is $Sample$-$wise$ SPSC, $Feature$-$wise$ SPSC and $Element$-$wise$ SPSC.
The effectiveness of proposed SPSC was demonstrated by experiments on real image dataset and it with noise. 
The proposed method shows its advantage over current SC methods on more accurately approximating the ground truth clustering, and we will evaluate it on other tasks, such as classification and recommendation. 

%\section{Acknowledgments}
\clearpage
\bibliography{ref_AAAI}
\bibliographystyle{aaai}
\end{document}